\documentclass[conference, letterpaper]{IEEEtran}
\IEEEoverridecommandlockouts
\ifCLASSINFOpdf
\else
\fi
\usepackage[cmex10]{amsmath}
\ifCLASSINFOpdf
   \usepackage[pdftex]{graphicx}
\else
\fi

\usepackage[cmex10]{amsmath}
\usepackage{color}
\usepackage{fancyhdr}
\usepackage[caption=false,font=footnotesize]{subfig}

\usepackage{array}

\usepackage[compress,sort]{cite}

\renewcommand{\thispagestyle}[2]{} 

\begin{document}

%
\title{Improved Training for Self Training by Confidence Assessments}

\author{
\IEEEauthorblockN{Dor Bank*}
\IEEEauthorblockA{Tel Aviv University\\
Tel Aviv, Israel\\
Email: dorbank@gmail.com}
\and

\IEEEauthorblockN{Daniel Greenfeld*}
\IEEEauthorblockA{Weizmann Institute of Science\\
Rehovot, Israel\\
Email: danielgreenfeld3@gmail.com}
\and
\IEEEauthorblockN{Gal Hyams*}
\IEEEauthorblockA{Tel Aviv University\\
Tel-Aviv, Israel\\
Email: gal.hyams@cs.tau.ac.il}
}


%


\maketitle

\begin{abstract}

It is well known that for some tasks, labeled data sets may be hard to gather. Self-training, or pseudo-labeling, tackles the problem of having insufficient training data. In the self-training scheme, the classifier is first trained on a limited, labeled dataset, and after that, it is trained on an additional, unlabeled dataset, using its own predictions as labels, provided those predictions are made with high enough confidence. Using credible interval based on MC-dropout as a confidence measure, the proposed method is able to gain substantially better results comparing to several other pseudo-labeling methods and out-performs the former state-of-the-art pseudo-labeling technique by 7 $\%$ on the MNIST data-set. In addition to learning from large and static unlabeled datasets, the suggested approach may be more suitable than others as an online learning method where the classifier keeps getting new unlabeled data. The approach may be also applicable in the recent method of pseudo-gradients for training long sequential neural networks.
\end{abstract}

\footnotetext[1]{*equally contributed authors, writers are presented by the alphabeth order.}
\footnotetext[2]{Acknowledgements: This study was supported in part by fellowships from the Edmond J. Safra Center for Bioinformatics at Tel-Aviv University and from The Manna Center for Food Safety and Security at Tel-Aviv University.}
\begin{IEEEkeywords}
Semi-supervised learning; Self-Training; Limited training set; MNIST; Image classification
\end{IEEEkeywords}

\section{Introduction}
In the semi-supervised learning scheme, both labeled and unlabeled data are being used to train a classifier. This is especially appealing when labeled data is very limited but unlabeled data is abundant, which is often the case when labeling new data is expensive and sufficient labeled data is not yet easy to find. Such tasks include semantic segmentation, sentence stressing, video labeling and more.
A very common example of limited labeled data but practically unlimited unlabeled data is the on-line stage of a classifier, in which the training data is fully exploited, but unlabeled test data keeps coming.
From a practical standpoint, self-training is one of the most simple approaches that can be utilized in those cases. 
 In this setting, after finishing the supervised training stage of a classifier, it is possible to continue the learning process of the classifier on new unlabeled data, which may be the on-line unlabeled test samples it is asked to classify. Whenever the classifier encounters a sample on which the certainty that the classification is correct is high, this sample can be used as a training example along with that prediction as replacement for a label.
 The crucial question is how can the self-training classifier decide on which of the self-labeled samples it should train on. In other words - when should the predictions of the not yet fully trained classifier be trusted?
 In this work, different methods for training a self-training classifier are suggested and their utilities are analyzed. The suggested techniques can be easily implemented on top of any boosting and data augmentation methods, improving the obtained results.\\
 The main contributions of this work are: (1) demonstrating how to use the self-training method in the most effective manner. Specifically, using the algorithm suggested here, state-of-the-art results for self-training on MNIST were achieved, improving the former state-of-the-art results by 7 $\%$. (2) Suggesting an empirical limitation of the self-training method, including an empirical lower bound on the preliminary success rate and data set size when implementing the self-training or C-EM method on multi-class classification image tasks.\\
The remainder of the paper is organized as follows: a description of previous works on semi-supervised and specifically self training classification is brought in the next section. In the \textit{Methods} section a detailed description of the techniques for deciding the trustworthiness of the classifier on unlabeled samples is given. In the \textit{Experiments} section, a comparison between the results of the approaches described on methods is presented. The contribution and importance of this article is further discussed in \textit{Discussion}. Finally, future works are suggested in the last section.

\section{Previous Works}

\subsubsection{Self-training}
The approach of self-training was first presented by Nigam \textit{et. al.} \cite{nigam2000analyzing} and it was shown that it can be interpreted as an instance of the Classification Expectation Maximization algorithm \cite{Amini:2002:SLR:3000905.3000988}.
Implementing the self-training method harnessing Denoising Auto-Encoder and Dropout \cite{hinton2012improving} on the visible layer in order to avoid over-fitting to the training set \cite{lee2013pseudo} have achieved the best known results for the self-training method on MNIST data-set so far.
\subsubsection{Semi Supervised MNIST Classification}
 Additional approaches for the semi-supervised task, demonstrated for image classification on MNIST:\\
 Generative Adversarial Networks \cite{salimans2016improved} have been successfully used in order to achieve the state-of-the-art results for semi-supervised MNIST classification, based on labeled dataset containing ten samples for each of the ten classes. On the other hand, such balanced labeled set is not always available, and training a GAN well may be difficult for a lot of tasks and datasets. \\ The unsupervised Ladder Networks technique \cite{valpola2015neural} was successfully harnessed for semi-supervised classification \cite{NIPS2015_5947} with success on the MNIST task as well. Implementing this method is not trivial for a variety of tasks.\\
 Another successful technique for contending with insufficient labeled data is the augmentation method, usually used for vision tasks.  We note that the method examined in this work can be used on top of augmentation even after the latter was exploited to its fullest; in addition, the method examined here can be efficiently used for multiclass classification tasks outside the area of computer-vision.  
 
\section{Methods}
As stated before, the proposed training is done as follows: the model is first trained on some labeled dataset, and then the model is trained on unlabeled data using its own predictions as ground-truth, whenever a confidence condition is met. In classification tasks, the standard method for extracting a confidence-measure is by looking at the soft-max layer probabilities. Unfortunately, networks tend to be over-confident in that sense, rendering those probabilities not informative enough. We therefore applied maximal entropy regularization as recently suggested in \cite{pereyra2017regularizing}, which penalizes the network whenever the soft-max probabilities are too concentrated in one class.
With that said, even if we know how to measure a networks confidence in its predictions, a crucial challenge remains - setting a confidence-threshold by which to decide whether to trust those predictions or not. The trade-off is clear - a low threshold will result in a high false-positive (\textit{FP}) rate which will cause the network to train on wrong samples; a high threshold will result in a low true-positive (\textit{TP}) rate which will mean that not enough additional data is obtained to make a difference. Furthermore, looking at the soft-max probabilities does not exactly yield a desired confidence measure. Those probabilities represent the network's best guess, but the quantity of interest here is to what extent is this guess reliable. We therefore turned to additional methods that help asses if a prediction is trustworthy: 1. Using MC-dropout as another way to represent a networks uncertainty \cite{gal2015dropout} by running the same network multiple times on a given sample, each run sending to zero a random sample from the hidden units of the network, thus obtaining a distribution over the network's predictions  and 2. Bagging of two networks - even when the networks have exactly the same architecture, the random initialization of the weights and the fact that the training is stochastic due to dropout layers are enough to ensure that the networks will yield different results on the borderline cases. That is, if they agree on a prediction - the chances of it being correct are much greater. \\
Concretely, the following methods were examined as ways to determine when a prediction is likely to be correct:

\subsection{Soft-max threshold}
In this method, a hard coded threshold is compared against the highest soft-max result. It is assigned as a hyper-parameter, which the user should provide. It can be assigned differently for each class, and intuitively it requires prior knowledge on the unlabeled data.

\subsection{Ensemble consensus}
A different approach, would be to ask for a vote from several classifiers. 
For convex models, this approach is obviously meaningless. For neural networks, it fits great, even when the networks has the same architecture, since each network is randomly initialized to different values. The downside of using an ensemble of classifiers is that it requires storing and training several networks, which is not applicable for most tasks.

\subsection{Dropout consensus} 
A more tractable version of such agreement test is to run one network several times with dropout $<1$ value, and check the consistency of the predictions as a proxy to an agreement test between several classifiers. Note that while a Majority-Vote was shown to be efficient \cite{kuncheva2003limits}, its efficiency is maximal when the voters are negatively depended on each other - which is not the case here. In addition, in order to minimize false positives, consensus vote is more effective.

\subsection{Several outputs expectation}
By using dropout, the number of outputs which is applicable can be quite large. Therefore, an approximation for the "true" output probability can be calculated empirically.

\subsection{Credible interval of outputs}
The downside of the agreement based methods, is that it doesn't take into account the distribution of the output probabilities. For example, suppose using $90\%$ as the threshold, and for a given sample getting 5 outputs of $95\%,95\%,95\%,95\%,20\%$. In this case, the sample won't be taken into account because their average is less than the threshold, even though the fifth output is probably an outlier. 
 In order to increase the number of unlabeled samples the model would be trained on while keeping a low false positive rate, a confidence interval on the prediction's probabilities was examined as well.
 Usually, confidence interval is used as a mean to estimate a parameter, when considered as a random variable - under the Bayesian paradigm. The output of the network is an estimation of the probability that the sample belongs to each of the examined classes. From these, it is possible to look on the network's outcomes as parameters for the  Multinomial distribution and construct a confidence interval for their values. From the central limit theorem, the distance of the empirical average of each parameter from the true parameter is distributed approximately like normal distribution. Now, it is possible to use the t-distribution in order to construct the confidence interval for the average of those parameters, as the true variances of the averages are not known. Since the average is an unbiased estimator to the expectation and the parameter equals to its expectation in the Multinomial distribution, these intervals are valid for the estimation of the probabilities that the sample belongs to each of the classes.\\

\section{Experiments}
The experiments were done on the MNIST dataset, which includes 50,000 training samples, and 10,000 test samples.
The training samples were split into 2 groups: labeled training data, and unlabeled data.
Different sizes of training data were tested: 100, 150, 300, and 1000.
The test data remained the same for comparison with other benchmarks.
The network's architecture is 2 convolutional layers, 2 fully connected layers, and a soft-max layer. For sampling, MC-Dropout[23] was used for achieving MAP estimator of the certainty[5-7].
When using only one network, the regularization coefficient $\lambda$ was decreased during training from one to zero, as suggested in \cite{grandvalet2005semi}. Using positive value of 1 for $\lambda$, punishing over confidence, and gradually decreasing this value all the way to -1, aiming to direct the classifier to be more sure in its predictions gave similar results to decreasing the $\lambda$ only to 0. Dropout value of $50\%$ was used for both the training stages.
The training process consisted of 40 epochs on the training data, and after that, training on the unlabeled data, where after each epoch, another epoch was done with the original training data. The training was done with one Tesla K80 GPU with 11G memory.

\subsubsection{Soft-max threshold}
The confidence threshold of the soft-max score was searched for empirically. Unfortunately, even when using entropy regularization - with a positive coefficient \cite{pereyra2017regularizing} or a negative coefficient \cite{grandvalet2005semi} and a very high threshold (up to 0.999), the method is resulted in too many false positives, causing it to fail.

\subsubsection{Ensemble consensus}
Ensemble of classifiers is a natural approach for lowering the rate of the false-positives predictions. Implementation of this technique for the image classification task was done by training two networks with different parameters. A sample would be taken for the second-stage training if the two networks have agreed on its prediction. Harsher criterion was suggested as well, including a constraint that each of the voters predictions will be higher then a certain probability. Even when using a confidence value of $50\%$, only a scarce set of examples have achieved the criteria. The first network was train with learning rate of $10^{-6}$ and with regularization parameter of 1, while the second with $10^-6$ and 0.1 respectfully. Results can be seen in \textit{Table 1} and \textit{Table 2}. 

 \subsubsection{Dropout consensus}
  In order to minimize even further the false positive rate, an additional constraint was enforced - the soft-max probability of the predicted class should be above a threshold of $95\%$. 
 This method was examined twice: once with a consensus of 80 voters, and once with 25 voters. Results are presented in \textit{table 3}.

 \subsubsection{Dropout credible interval}
 For applying this method, the classifier learned from samples
for which the lower value of the interval was higher then the threshold.
  As a trade-off between the running time and accuracy, 80 MC-Dropout iterations were used. The credible interval was the common $95\%$ one.\\ For improving the results, the threshold of the lower value was initially set to 0.98 and then slowly decreased all the way to 0.9. By that, the network first trained on the example it was most sure of and only later handling  with the less obvious examples. 
  
\subsection{Results}
This section presents the labeled data size, the initial test accuracy after training on merely the labeled data only, the test accuracy after the entire training session (\textit{best acc}), the true positive rate among the positives (\textit{TP} ) and the Positive rate (\textit{P}) which is the rate of the unlabeled data that was used in the second training phase.

Results for using consensus with two networks:

\begin{table}
\begin{center}
\begin{tabular}{|c|c|c|c|c|}\hline
training size & basic acc  & best acc & TP & P \\[1 ex] \hline\hline
100  & 0.7  &  fail  & 0.47 & 0.92\\
150  & 0.74 &  0.8  & 0.81 & 0.83\\
300  & 0.87  & 0.95 & 0.95 & 0.99\\
1000 & 0.90  & 0.96 & 0.97 & 0.98 \\ \hline
\end{tabular}
\end{center}
\caption{Consensus two networks - with entropy minimization}
\label{table:1}
\end{table}

\begin{table}
\begin{center}
\begin{tabular}{|c|c|c|c|c|}\hline
training size & basic acc  & best acc & TP & P \\ \hline\hline
100  & 0.7  &  fail  & 0.75 & 0.84\\
150  & 0.81 &  0.86  & 0.93 & 0.88\\
300  & 0.85  & 0.95 & 0.96 & 0.98\\
1000 & 0.92  & 0.97 & 0.95 & 0.98 \\ \hline
\end{tabular}
\end{center}
\caption{Consensus of two networks - with entropy maximization}
\label{table:2}
\end{table}

In \textit{table 1} and \textit{table 2}, the results for self training based on the consensus of two networks with two different regularization are presented. \textit{Entropy minimization} regulator punishes the network when the classifier is not sure enough in it's results, making the predictions to be more polarized. This may help the classifier notice the underling separation of the data to different classes, even when it knows little on each of the classes. From the other side, \textit{Entropy maximization} punished on over-confidence, making the classifier predict with high probability only when it is absolutely sure in the prediction. 
Notice that the results are comparable, regardless of the entropy regularization. This is somewhat expected since this regularization is applied in order to influence the networks own certainty about its predictions. In this approach the networks confidence was not taken into account when deciding whether or not to trust the predictions. It was simply examined if the two networks agreed. Another thing to note is that label set of only 100 samples is not enough for this approach; more specifically, an initial accuracy of  $70\%$  was to low for the technique. With that said, even a slight improvement (to an accuracy of  0.74) lead to a much better \textit{TP} rate and therefore to an improvement in the second stag0e of the training. Finally, a well known phenomena can be observed here - reaching a 0.95-0.97 accuracy was relatively easy and happened even when the initial accuracy was only at 0.85, but an accuracy of 0.99 remained outside of our reach even when starting with an initial 0.92 accuracy. A more accurate decision criteria is needed in order to break this glass ceiling.

 \begin{table}
 \begin{center}
\begin{tabular}{|c|c|c|c|c|}\hline
training size & basic acc  & best acc & TP & P \\ \hline\hline
100  & 0.68  &  fail  & 0.53 & 0.2\\
150  & 0.798 &  fail  & 0.61 & 0.2\\
300  & 0.85  & 0.8825 & 0.86 & 0.2\\
1000 & 0.89  & 0.9 & 0.94 & 0.2 \\ \hline
\end{tabular}
\end{center}
\caption{Consensus among 80 Dropout voters}
\label{table:3}
\end{table}

In table 4, the results when running the predictions with 25 Dropout voters are presented. A lower number of voters increase the likelihood of agreement among the voters.

\begin{table}
\begin{center}
\begin{tabular}{|c|c|c|c|c|}\hline
training size & basic acc  & best acc & TP & P \\ \hline\hline
100 & 0.68 & fail & 0.1 & 0.82\\
150 & 0.75 & fail & 0.59 & 0.75 \\
300 & 0.84 & 0.91 & 0.9 & 0.75\\
1000 & 0.91 & 0.93 & 0.95& 0.75 \\ \hline
\end{tabular}
\end{center}
\caption{Consensus among 25 Dropout voters}
\label{table:4}
\end{table}

It is possible to see that the self training approach based on the \textit{Dropout-consensus} is quite sensitive to the initial training stage and could not maintain satisfying results from a very small labeled data set.
For the \textit{Dropout Confidence Interval} method, satisfying results were achieved even with training size=100. Therefore, there was no need for larger labeled sets. Based on this training set, the initial accuracy was  $73\%$; after 100 epochs finished with $95.94\%$ (
\textit{TP}=$99\%$, \textit{P}=$87.6\%$). Applying the deteriorating confidence threshold (together with gradually decrease in the regularization coefficient, from 1 all the way down to $-0.5$) yield slightly better accuracy of $96.58\%$ with (\textit{TP}=$98.7\%$, \textit{P}=$92.8\%$). The higher percentage of true positive can be explained by the lower threshold at the end of the training stage.

 \begin{table}
 \begin{center}
\begin{tabular}{|c|c|c|c|c|}\hline
Method & error rate \\ \hline\hline
improved GAN*  & 0.86 \cite{salimans2016improved} \\
Ladder  & 1.06 \cite{NIPS2015_5947} \\
AtlasRBF & 8.10 \cite{pitelis2014semi} \\
pseudo-label & 10.49 \cite{lee2013pseudo}\\
self training with CI  & 4.06  \\
self training with CI and dynamic confidence & 3.42\\\hline 
\end{tabular}
\end{center}
\caption{Comparative with other semi-supervised classification methods on the MNIST data-set}
\label{table:5}
\end{table}
Table 5 contains the comparison with other semi-supervised techniques. This suggested technique outperform the previous successful implementation of the self-train method on the MNIST set, which used auto-encoders and dropout. Our approach is second only to much more sophisticated methods, and can be used simply for any semi-supervised classification task.\\
When considering the confidence interval, the classifier becomes much more robust. Interestingly, when the decision was based on the interval's lower value, a recovering mechanism took place as the epochs on the data progressed: During the training procedure on the smallest datasets, the accuracy of the network initially got worse than it was after the initial training on the labeled set. Notwithstanding, as the training continued, the accuracy improved, achieving much better results - and passing the best results known so far using the self-training technique. In the experiments that was conducted based on 100 labeled examples, the test accuracy first went down to approximately $15\%$ after a few epochs on the data, only to come back up to its final presented result.
In addition, the amount of samples the classifier learned from was almost monotonously increasing as the iterations on the set progressed. This implies that the network did always converge towards a local minimum.
To explain why superior results where obtained when the confidence interval was considered, the true-positives (\textit{"TP"}) and total positives (\textit{"P"}) rates is examined: Obviously, each sample that is labeled as Positive under the \textit{consensus $\&$ above threshold} criteria, will be labeled as positive under the \textit{Dropout Confidence Interval} criteria. On the other hand, the \textit{Dropout Confidence Interval} criteria classifies much more samples as Positive, and judging by the \textit{TP} percentage, most of them are \textit{TP}. Therefore, compared to the other examined criteria, the \textit{Dropout Confidence Interval} manage to train on more samples (while keeping a low rate of false positives), which increases the generalization ability of the classifier. When using an initial random labeled training set of 80, the initial rate of successful predictions of the classifier was $67\%$. In this case, the classifier failed to recover from the low \textit{TP} rate. This suggests a lower bound on the initial accuracy when using the self-training approach on semi-supervised vision tasks.
\section{Discussion}
The work brought here examines the intuitive and flexible approach of self-training as a semi-supervised approach for computer vision tasks. 
The best self-training algorithm presented here out-preformed the former state-of-the-art self-training technique by 7 $\%$ on the MNIST data-set. In addition, this algorithm is simple to implement on various networks for variable datasets and is likely to work for when done right and when the initial success rate is satisfactory. An interesting usage may be using the self-training algorithm described here for recently-popular pseudo-gradient approach, aiming to train recurrent neural networks (RNN) on long sequences. In this approach, the long sequence is first divided to shorter sequences. Then, the RNN is trained on each of these short sequences without waiting to the gradient from the loss on the successive sequences. Instead, the RNN 'guesses' these gradients, assigns its guess as a pseudo-gradient and continues the training based on this pseudo gradient.

\section{Future work}
Following the conclusions, much work can be done to follow up.
In addition to the basic MNIST classification task, Self-training for semantic segmentation was examined as well.
Unfortunately, the segmentation network did not achieved the minimal success rates after the initial supervised training on a limited labeled set, resulting in failing to show positive results for the semi-supervised semantic segmentation task with limited data set.\\
It is reasonable that the self-trained method would work for semantic segmentation as well, given a sufficient preliminary accuracy. For achieving this, it's advised to first train the model over all of the available labeled dataset, and then continue the self-training stage over another data-set similar to the labeled one. 
Focusing on the segmentation task, the confidence threshold should probably not be a fixed number. It should vary for each pixel, based on nearby pixels, and should change based on classes appearances.\\
Interesting work can be done by harnessing the recent advances in low-shot visual recognition \cite{hariharan2016low} to the challenging semi-supervised semantic segmentation task. In addition, harnessing sophisticated techniques for coping with biased prediction, especially in segmentation\cite{bulo2017loss}, can be of great use when trying to learn from a minimal random data-set.\\
Another interesting work that can be done using the self-training method is unsupervised classification, similarly to K-means. The advantage of the self-training method over k-means is its wider approach, which can come into practice by considering higher statistical moments than just the first one - in C-EM, for example, and more generally, by learning the underling representation of the class in an implicit manner, which can eventually be more accurate. Maintaining effective choosing of class seeds, as described in k-means++ \cite{arthur2007k} and its following works, can result in an adequate unsupervised discriminator, especially when using the Entropy minimization regularization \cite{grandvalet2005semi}.
Finally, one last thing to additionally explore is the performance of the network when predictions in the final stage of testing the model, after the semi-supervised training was complete, are made in the same manner as was made during the semi-supervised training. For example, by using the MC-Dropout.



%

\bibliographystyle{plain}
\bibliography{for_arxiv}

\end{document}